\documentclass{article}

\usepackage{nips_2018}

\usepackage{algorithm}
\usepackage{amsfonts}		% blackboard math symbols
\usepackage{amsmath}
\usepackage{booktabs}		% professional-quality tables
\usepackage{color}
\usepackage{comment}
\usepackage{etoolbox}
\usepackage[T1]{fontenc}	% use 8-bit T1 fonts
\usepackage{hyperref}		% hyperlinks
\usepackage[utf8]{inputenc}	% allow utf-8 input
\usepackage{microtype}		% microtypography
\usepackage{nicefrac}		% compact symbols for 1/2, etc.
\usepackage{url}			% simple URL typesetting
\usepackage{wrapfig}

\DeclareMathOperator*{\argmax}{arg~max}

\usepackage{algorithmicx}
\usepackage{algpseudocode}
\usepackage{graphicx}
\usepackage{subcaption}

\newbool{inccomment}
\setbool{inccomment}{true}
\newcommand{\XX}[1]{\ifbool{inccomment}{{\color{magenta} #1}}{}}
\newcommand{\CT}[1]{\ifbool{inccomment}{{\color{magenta}CT\@: #1}}{}}
\newcommand{\NT}[1]{\ifbool{inccomment}{{\color{blue}NT\@: #1}}{}}
\newcommand{\TD}[1]{\ifbool{inccomment}{{\color{orange}#1}}{}}
\newcommand{\FN}[1]{\ifbool{inccomment}{{\color{OliveGreen}#1}}{}}
\newcommand{\GR}[1]{\ifbool{inccomment}{{\color{Tan}#1}}{}}
\newcommand{\LD}{\ifbool{inccomment}{{\color{magenta}\\============================================\\}}}
\newcommand{\RF}{\ifbool{inccomment}{{\color{green}~[R]}}}

\newcommand{\roma}[1]{\uppercase\expandafter{\romannumeral #1\relax}}

\title{Demystifying Neural Network Filter Pruning}

\author{
	Zhuwei Qin$^1$, Fuxun Yu$^2$, Chenchen Liu$^3$ Xiang Chen$^4$ \\
	$^{1,2,4}$Department of Electrical Computer Engineering, George Mason University, Fairfax, VA 22030 \\
	$^{3}$Department of Electrical Computer Engineering, Clarkson University, Potsdam, NY 13699 \\
	\texttt{zqin@gmu.edu$^1$, fyu2@gmu.edu$^2$, chliu@clarkson.edu$^3$, xchen26@gmu.edu$^4$}
}

\begin{document}
\maketitle
\begin{abstract}
Based on filter magnitude ranking (e.g. $\ell_1$ norm), conventional filter pruning methods for Convolutional Neural Networks (CNNs) have been proved with great effectiveness in computation load reduction.   
	Although effective, these methods are rarely analyzed in a perspective of filter functionality.
In this work, we explore the filter pruning and the retraining through qualitative filter functionality interpretation.
% a perspective of filter functionality.
	We find that the filter magnitude based method fails to eliminate the filters with repetitive functionality.
	And the retraining phase is actually used to reconstruct the remained filters for functionality compensation for the wrongly-pruned critical filters.
	%The nature of retraining process is dramatically reconstructing the network rather than filter functionality tuning.
With a proposed functionality-oriented pruning method, we further testify that, by precisely addressing the filter functionality redundancy, a CNN can be pruned without considerable accuracy drop, and the retraining phase is unnecessary.
% slightly accuracy drop redundancy without retraininout cg.
\end{abstract}

%A widely accepted assumption is that the convolutional filters with smaller magnitude (e.g. $\ell_1$ norm) can be pruned since they play less important roles at the inference time.
\vspace{-2mm}
\section{Introduction}
The great success of CNN is benefited from its complex algorithm and architecture at a cost of intensive computation load.
	Therefore, many CNN filter pruning works have been proposed to alleviate this issue [1-7].
%The filter pruning works are mostly based on a quantitative evaluation that the convolutional filters with smaller magnitude are less important.
%moticvation
While these works demonstrated expected performance, most of them are merely based on quantitative ranking of the filters' magnitude.
	However, how the magnitude ranking really reflect the filters' functionality and contribution to the classification still remains a lack of research.
	%The impact of magnitude based filter pruning on filter functionality is rarely to be considered.
Therefore, in this work, we utilized the filter visualization technique to interpret one of the most representative magnitude ranking based filter pruning method (i.e. $\ell_1$ norm [1]), as well as the retraining process in a perspective of filter functionality.
	From the analysis, we find that: 
\vspace{-1mm}
\begin{itemize}
	\item The magnitude ranking based filter pruning method fails to select filters with functionality redundancy, resulting in inevitable accuracy drop;
	%\vspace{-0.5mm}
	\item Filters suffer from significant functionality changes during retraining phase, which indicates that the magnitude ranking based filter pruning method may defect certain critical filter functionality and requires filter reconstruction for compensation;
	% prune magnitude-wise small but functionality-wise critical filters. 
	%The remaining filters should be reconstructed to compensate the missing functionality;
	%\vspace{-0.5mm}
	\item With functionality-oriented pruning method, a CNN can actually be pruned without considerable accuracy drop and the retraining phase is relatively unnecessary.
	% slightly accuracy drop by our proposed functionality-oriented filter pruning method without retraining, which precisely reduces the filter functional redundancy.
\end{itemize}
\vspace{-1mm}

% \item We propose a functionality-oriented filter pruning method to further testify that the network can be 		pruned with slightly accuracy drop without retraining by precisely reducing the filter functional redundancy.
%  The neural network can be pruned with slightly accuracy drop without retraining by our proposed functionality-oriented filter pruning method, which precisely reduce the filter functional redundancy.
%	\item The filter magnitude based pruning methods prune magnitude-wise small but functional-wise critical filters.
	% The remaining filter should be reconstructed by retraining to compensate the missing functionality;

	% the filter magnitude based pruning methods defects the composition of filter functionality.
\vspace{-3mm}
\section{Analyzing Magnitude Ranking based Filter Pruning \hspace{6cm} through Filter Functionality Interpretation}

\begin{figure}[t]
	\centering
	\includegraphics[width=5.5in]{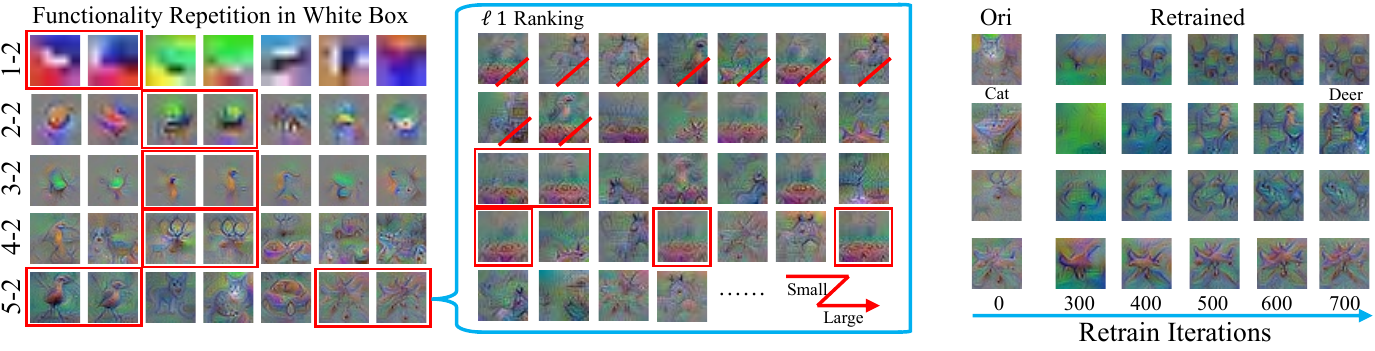}
	\vspace{-4mm}
	\caption*{\hspace{-2mm}(a) Visualized Patterns \hspace{20mm} (b) Pruned Filters \hspace{20mm}(c) Network Retraining}
	\vspace{-1mm}
	\caption{Case study of the $\ell_1$ based filter pruning on the Conv5\_2 of VGG-16 and network retraining interpretation based on filter functionality.}
	\vspace{-3mm}
	\label{fig:pruning_visua}
\end{figure}

In this section, we utilize a well-established CNN visualization technique [8] -- Activation Maximization (AM) to interpret the filter functionality.
% The AM visualizes a single filter's functional preference for feature extraction by an input pattern that can maximize the filter's activation.
	In the CNN visualization analysis, the filter functionality is usually defined as the feature extraction preference, which can be represented by a synthesized input image that causes the highest feature map activation for a filter during the convolution process.
Mathematically, the visualization process can be formulated as:
\small
	\begin{equation}
		V(F^l_i)=\argmax_{X} {A^l_i(X),
		\hspace{0.6cm} X \leftarrow X} + \eta \cdot \frac{\partial A^l_i(X)}{ \partial X},
		\label{eq:am}
	\end{equation}
\normalsize
	where $A^l_i(X)$ is the activation of filter $F_i^l$ from an input image $X$, $\eta$ is the gradient ascent step size.
	With $X$ initialized as an input image of random noises, each pixel of this input image is iteratively changed along the $\partial A^l_i(X)$/$\partial X$ increment direction to achieve the highest activation.
	Eventually, $X$ demonstrates a specific visualized pattern $V(F^l_i)$, which contains the filter's most sensitive input features with certain semantics, and represents the filter's functional preference for feature extraction.

For preliminary demonstration, an VGG-16 model trained on the CIFAR-10 dataset is adopted [1]:
(1) Fig.~\ref{fig:pruning_visua} (a) demonstrates filters' visualized patterns from different layers for functionality interpretation analysis.
	From Fig.~\ref{fig:pruning_visua} (a), we find many similar patterns inside each layer (denoted in red blocks).
	The filters with similar functionality may repetitively extract the same feature and introduce significant network redundancy.
	%These similarities indicate the functionality repetition of the convolutional filters, which could introduce considerable redundancy into CNNs.
(2) Fig.~\ref{fig:pruning_visua} (b) demonstrates a case study of a pruned layer (Conv5\_2) based on the magnitude ranking of $\ell_1$ norm.
	As shown in the figure, the filters' visualized patterns are ranked by the $\ell_1$ norm in an ascending order, where the pruned filters are marked by red slashes.	
We can observe that the $\ell_1$ norm based pruning preserves all the filters with high magnitudes, even significant functionality repetition exists among those filters.
	Therefore, the $\ell_1$ norm based pruning fails to address the functionality redundancy in the model.
	%While, the filters with similar functionality may repetitively extract the features and introduce significant network redundancy.

%to eliminate the filters with repetitive functionality to choose filters with functional redundancy.
According to the functionality interpretation, the filters with small magnitudes could also demonstrate distinct feature extraction preferences, and contribute to the feature extraction integrity and diversity.
	The magnitude ranking based pruning method may overlook the significance of small filters and defect the information retrieval for critical features, resulting in inevitable accuracy drop.
% play important role during inference time.
%Thus, the $\ell_1$ norm based method will result in an inevitable accuracy drop.

\section{Filter Functionality Transition During Retraining Phase}

In most neural network pruning works, the retraining phase is a mandatory operation for maintaining the accuracy performance.
%to compensate the accuracy drop.
	However, the iteratively pruning and retraining operation also introduce computation cost.
In this section, we analyze the mechanism and necessity of the retraining phase.
%impact of the retraining process on the filter functionality and revel the mechanism of the retraining process on compensating the accuracy drop.
	%is qualitatively evaluated by the AM. 
	%which demonstrates the filter functionality transition during network retraining.
\begin{figure}[b]
	\centering
	\includegraphics[width=5.5in]{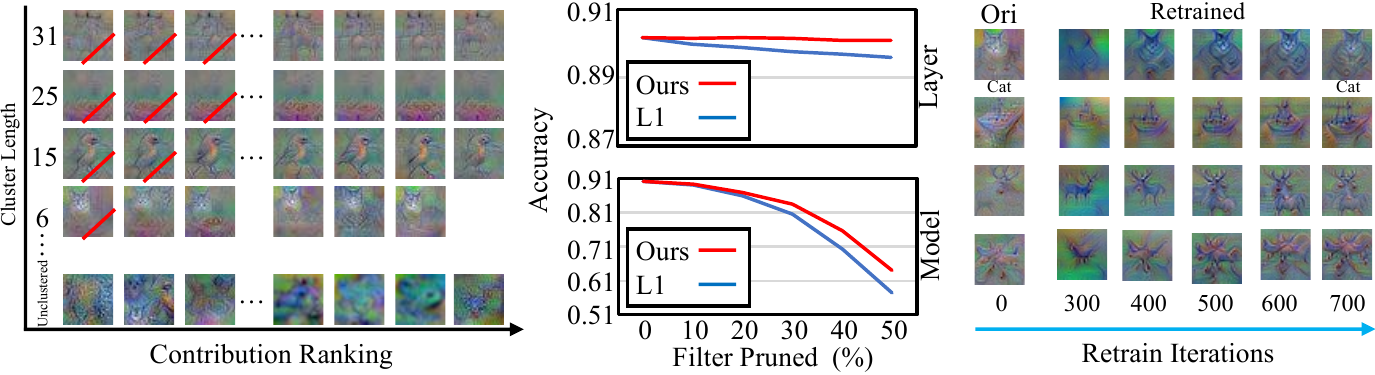}
	\caption*{\hspace{-4mm}(a) Functionality-Oriented Filter Pruning \hspace{6mm} (b) Accuracy Drop \hspace{10mm} (c) Network Retraining \hspace{10mm}}
	\caption{Case study of the filter functionality-oriented filter pruning on the Conv5\_1 of VGG-16.}
	\label{fig:cluster}
	\vspace{-5mm}
\end{figure}

As shown in Fig.~\ref{fig:pruning_visua} (c), we randomly select four filters that have not been pruned by the $\ell_1$ norm method. 
Then we use the visualization to revel the filters' original functionality and the functionality transition during different retraining iterations, e.g. every $100$ iterations. 
	From Fig.~\ref{fig:pruning_visua} (c), we can observe that: 
		(1) For most filters, the visualized patterns are dramatically changed by the $\ell_1$ morn based pruning method.
		For example, the content of the visualized patterns in the first row changes from a cat to a deer.
		% which indicates the retraining process of the $\ell_1$ based method reconstructs the filter functionality.
		Such changes indicate that the retraining phase eventually reconstruct the remaining filters' functionality to compensate the accuracy drop. 		
		(2) During the retraining iterations, the filters' functionality construction is gradually implemented, which indicates that a certain amount of retraining iterations and the corresponding computation cost are inevitable.
		% can be  The dramatic change has to be implemented by iterative retraining phases,Between every retraining iteration, the filter the  significantly changes functionality during different retraining stages.
		%(2) Eventually, the final visualized patterns (e.g. iteration 700) are significantly different from the original patterns.

		% With increasing retraining iterations, the visualized patterns are becoming more clear, which indicates the retraining process of the $\ell_1$ based method reconstructs the filter functionality, which therefore needs more retraining iterations to recover the model accuracy. 
	
Therefore, the nature of retraining process is to dramatically reconstruct the network rather than filter functionality fine-tuning.
	Based on the previous analysis, such reconstruction might be introduced by missing filters with significant functionality but small magnitude, and the remaining filters are reconstructed for functionality compensation.
	%Therefore, the retraining phase is necessary 
	%To compensate the missing functionality, 
%Hence, the $\ell_1$ based pruning method partially defects the original neural network's functionality composition.
%The remaining filter should be reconstructed by retraining to compensate the missing functionality, which balance the network functionality composition.
\section{Functionality-Oriented Filter Pruning}

Different from the $\ell_1$ norm based filter pruning method, we propose a  functionality-oriented filter pruning method, which are expected to precisely reduce the filter functionality redundancy.

Fig.~\ref{fig:cluster} (a) illustrates an intuitive example of our method in the layer Conv5\_1:
Each row represents one filter cluster, where the filters with similar functionalities are grouped by applying K-means analysis to the Euclidean distance of the visualized patterns. 
The last row shows the filters with extremely minimal similarity to each other, which are not considered for pruning due to their possible instinct functionality.
As each cluster contain different filter numbers, the filters are sorted by their contribution index $\gamma_{i}$, which is evaluated by the back-propagation gradients analysis:
\small
\vspace{-0.5mm}
\begin{equation}
	\medmuskip=-2mu
	\gamma_{i} = \frac{1}{N}\sum_{n=1}^{N} \left \| \frac{\partial Z}{\partial A_{i}(x_{n})} \right \|,
	\label{eq:contri}
	\vspace{-0.5mm}
\end{equation}
\normalsize
where the $Z$ and $A_{i}(x_{n})$ is the CNN output and filter $i$'s activation for each test image $n$ respectively.
Given certain filter pruning amount, the relative pruning rate for each cluster can be determined by cluster's volume size: the cluster with more repetitive filters will be pruned more aggressively. 
In each cluster, the filters with least contribution will be pruned first. 

\vspace{-1mm}
\begin{wrapfigure}{r}{1.7in}
  \vspace{-7.5mm}
  \captionsetup{justification=centering}
  \includegraphics[width=1.7in]{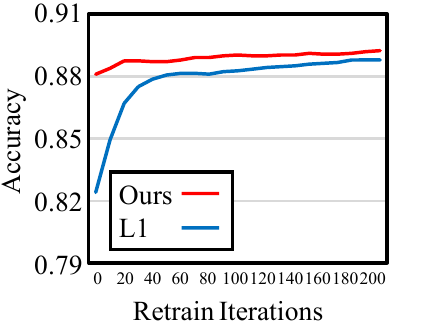}
  \caption{Pruned Model Accuracy Recovery}
  % \caption*{Figure 3: Pruned model accuracy recovery.}
  \label{fig:retrain}
  \vspace{-2mm}
\end{wrapfigure}

The aforementioned method can be applied to each convolutional layers concurrently to reduce the whole CNN functionality redundancy.
Fig.~\ref{fig:cluster} (b) shows the layer-wise (Conv5\_1) and model-wise CNN accuracy drop under different pruning ratio respectively.
We can see that our method has a slower accuracy drop compared with the $\ell_1$ norm based pruning method in both layer-wise and model-wise pruning.
In the Fig.~\ref{fig:cluster} (c), the retraining process of our method is also qualitatively evaluated.
We can observed that, regardless the retraining iterations, the remained filters' functionality remain unchanged.  
For example, the content of visualized pattern in the first row is still a cat at iteration 700, while the filter functionality is even clearer after retraining. 
That explains why our pruning method has less accuracy drop.

Meanwhile, we also quantitatively evaluate the retraining phase's effectiveness with the proposed method.
% the retraining process in terms of the model accuracy recovery.
  As shown in Fig.~\ref{fig:retrain}, the red line represents the pruned model accuracy recovery based on our proposed method whereas the blue line represents the $\ell_1$ norm based pruning method.
  We can see that the model pruned by our method demonstrates much less impact on the accuracy. 
  And the accuracy change is mainly introduced by filter fine-tuning, which also takes much less iteration numbers.
  %more quickly accuracy recovery.
  Therefore, with more interpretable and accurate repetitive filter identification and functionality-oriented pruning, the costly retraining phase becomes less necessary.

\section{Conclusion}
In this work, we interpret the magnitude filter pruning including retraining phase in a perspective of filter functionality. 
We show that the filter magnitude pruning method fails to choose filters with functional redundancy.
By further analyzing the functionality transition of remaining filters in the retraining phase, 
we revealed that the magnitude based pruning actually partially destructs original neural network’s functionality composition. 
The nature of retraining phase is dramatically network reconstruction rather than recover the filter functionality.
By contrast, our proposed functionality-oriented method demonstrated consistent filter functionality during retraining phase, indicating less harm to original network functional composition.

\newpage
\section*{References}
%Classic Filter Pruning
[1] H. Li, A. Kadav, I. Durdanovic, H. Samet, and H. P. Graf. (2017)
	Pruning Filters for Efficient ConvNets.
	in Proceedings of {\it the International Conference on Learning Representations} (ICLR),
	pp. 1-13.

[2] J.-H. Luo, J. Wu, and W. Lin. (2017)
	ThiNet: A Filter Level Pruning Method for Deep Neural Network Compression.
	in Proceedings of {\it the International Conference on Computer Vision} (ICCV),
	pp. 5058-5066.

[3] Y. He, X. Zhang, and J. Sun. (2017)
	Channel Pruning for Accelerating Very Deep Neural Networks.
	in Proceedings of {\it the International Conference on Computer Vision} (ICCV),
	pp. 1389-1397.

[4] H. Hu, R. Peng, YW. Tai, and CK. Tang. (2016) 
	Network trimming: A data-driven neuron pruning approach towards efficient deep architectures. 
	{\it arXiv:1607.03250}.

%compression

[5] S. Han, H. Mao, and W. J. Dally. (2014) 
	 Deep compression: Compressing deep neural networks with pruning, trained quantization and huffman coding.
	 {\it arXiv:1510.00149}.

[6] M. Jaderberg, A. Vedaldi, and A. Zisserman. (2014)
	 Speeding up convolutional neural networks with low rank expansions.
	 {\it arXiv:1405.3866}.

[7] W. Wen, C. Wu, Y. Wang, Y. Chen, and H. Li. (2016) 
	 Learning structured sparsity in deep neural networks.
	 in Proceedings of {\it the Advances in Neural Information Processing Systems} (NIPS),
	 pp. 2074-2082.

%Visualization

[8]  J. Yosinski, J. Clune, A. Nguyen, T. Fuchs, and H. Lipson. (2015) 
	 Understanding neural networks through deep visualization. 
	 {\it arXiv:1506.06579}.

\end{document}